\documentclass[10pt,twocolumn,letterpaper]{article}
\usepackage[pagenumbers]{cvpr} 

\usepackage{url}
\usepackage{graphicx}
\usepackage{amsmath}
\usepackage{amssymb}
\usepackage{booktabs} 
\usepackage{colortbl}
\definecolor{cvprblue}{rgb}{0.21,0.49,0.74}
\usepackage[pagebackref,breaklinks,colorlinks,allcolors=cvprblue]{hyperref}

\makeatletter
\def\@fnsymbol#1{%
   \ifcase#1\or
   \TextOrMath \textdagger \dagger\or
   \TextOrMath \textdaggerdbl \ddagger \or
   \TextOrMath \textsection  \mathsection\or
   \TextOrMath \textparagraph \mathparagraph\or
   \TextOrMath \textbardbl \|\or
   \TextOrMath {\textdagger\textdagger}{\dagger\dagger}\or
   \TextOrMath {\textdaggerdbl\textdaggerdbl}{\ddagger\ddagger}\else
   \@ctrerr \fi
}
\makeatother

\begin{document}
\title{VistaDream: Sampling multiview consistent images \\for single-view scene reconstruction}


\author{
Haiping Wang\textsuperscript{1}\ \ 
Yuan Liu\textsuperscript{2,3,}\thanks{Corresponding authors.}\ \ \ \ 
Ziwei Liu\textsuperscript{3}\ \ 
Wenping Wang\textsuperscript{4}\ \ 
Zhen Dong\textsuperscript{1,\textdagger}\ \ 
Bisheng Yang\textsuperscript{1}\\
\textsuperscript{1}Wuhan University\ \ 
\textsuperscript{2}Hong Kong University of Science and Technology\\
\textsuperscript{3}Nanyang Technological University \ \ 
\textsuperscript{4}Texas A\&M University\\
{\tt\small \{hpwang,dongzhenwhu,bshyang\}@whu.edu.cn}\ \ \ \ \ \ \\
{\tt\small yuanly@ust.hk}\ \ \  {\tt\small ziwei.liu@ntu.edu.sg}\ \ \ {\tt\small wenping@tamu.edu}\\
}

\twocolumn[{%
\renewcommand\twocolumn[1][]{#1}%
\maketitle
\vspace{-15pt}
\includegraphics[width=0.98\linewidth]{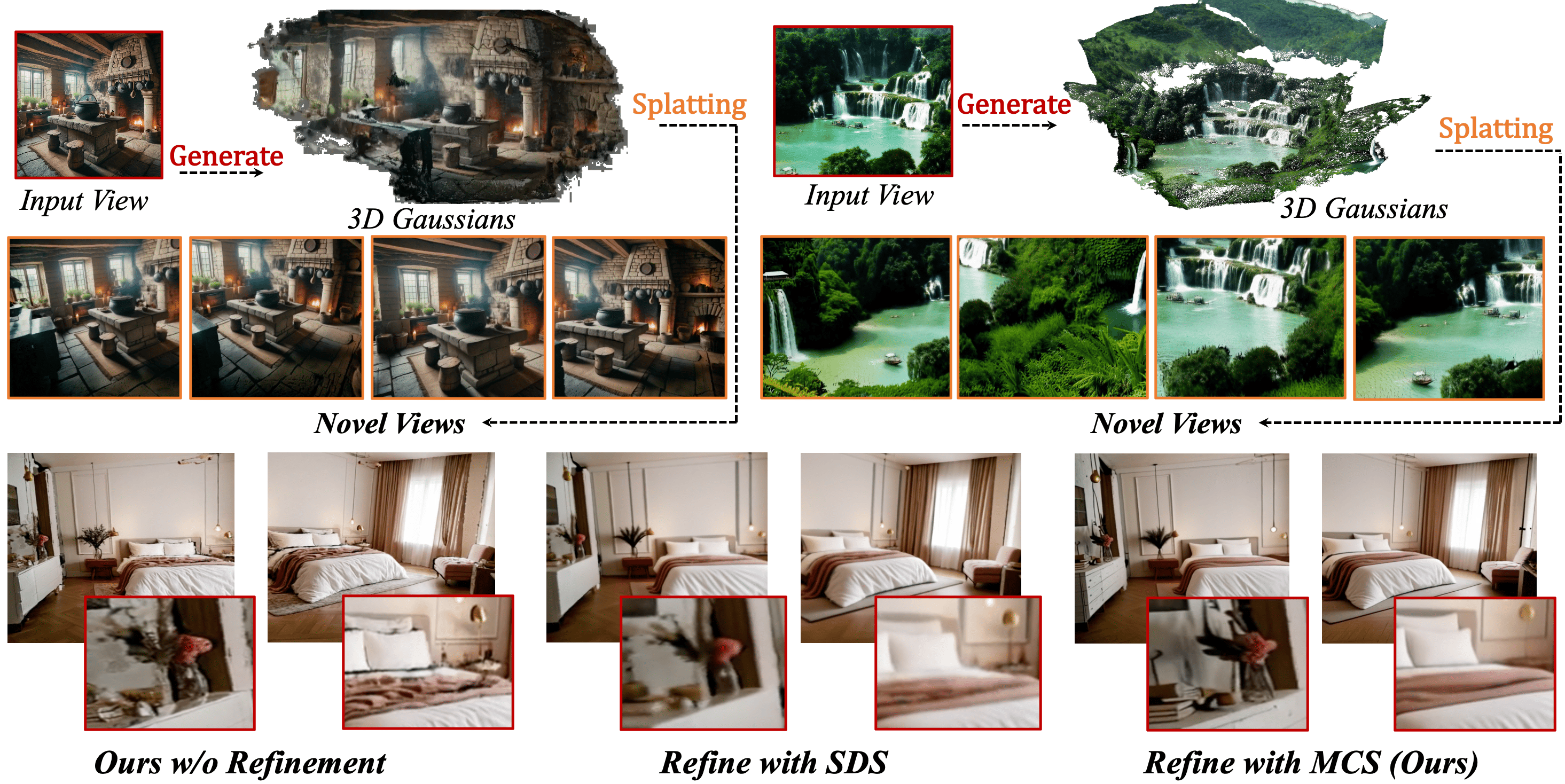}
\vspace{-5pt}
\captionof{figure}{\textit{Overview.} (Top) Given a single-view image of a scene, VistaDream reconstructs a 3D scene represented by 3D Gaussian Splatting (3DGS)~\citep{gs} for novel view synthesis.
(Bottom) The proposed Multiview Consistency Sampling (MCS) significantly improves scene quality and achieves better results compared to the commonly used Score Distillation Sampling (SDS)~\citep{sds}.}
\label{fig:teaser}
\vspace{10pt}
}]
{
  \renewcommand{\thefootnote}%
    {\fnsymbol{footnote}}
  \footnotetext[1]{Corresponding authors.}
}
\begin{abstract}

\vspace{-10pt}

In this paper, we propose VistaDream a novel framework to reconstruct a 3D scene from a single-view image. Recent diffusion models enable generating high-quality novel-view images from a single-view input image. Most existing methods only concentrate on building the consistency between the input image and the generated images while losing the consistency between the generated images. VistaDream addresses this problem by a two-stage pipeline. In the first stage, VistaDream begins with building a global coarse 3D scaffold by zooming out a little step with inpainted boundaries and an estimated depth map. Then, on this global scaffold, we use iterative diffusion-based RGB-D inpainting to generate novel-view images to inpaint the holes of the scaffold. In the second stage, we further enhance the consistency between the generated novel-view images by a novel training-free Multiview Consistency Sampling (MCS) that introduces multi-view consistency constraints in the reverse sampling process of diffusion models. Experimental results demonstrate that without training or fine-tuning existing diffusion models, VistaDream achieves consistent and high-quality novel view synthesis using just single-view images and outperforms baseline methods by a large margin. The code, videos, and interactive demos are available at \url{https://vistadream-project-page.github.io/}.

\end{abstract}

\vspace{-15pt}
\section{Introduction}

Reconstructing 3D scenes is a critical task in computer vision, robotics, and graphics. Traditionally, this has required multiple images from different viewpoints~\citep{colmap} or specialized hardware like RGBD scanners~\citep{kinectfusion} to capture both geometry and appearance. However, in real-world applications like AR/VR and robotics, we often only have access to a single-view image. Single-view 3D scene reconstruction is a highly challenging, ill-posed problem. Recent advances in diffusion models~\citep{ddpm} demonstrate strong capabilities in generating realistic images, offering promise for creating novel views from single images to aid in 3D reconstruction. The key challenge, however, is ensuring consistency across the generated views to produce coherent 3D scenes.

To tackle this challenge, prior works~\citep{tseng2023consistent,yu2023long,muller2024multidiff,wonderjourney,wonderworld} have primarily focused on enforcing consistency between the input single-view image and the generated novel views but struggle with ensuring consistency between the generated views themselves. Early approaches~\citep{tseng2023consistent,yu2023long} introduced techniques like epipolar line attention to aligning the input and generated views. More recent methods~\citep{genwarp,wonderjourney,wonderworld,realmdreamer} adopt a warp-and-inpaint approach, where they estimate depth, warp the image to a new viewpoint, inpaint it, and repeat this process iteratively to reconstruct the 3D scene. While promising, these methods still suffer from inconsistencies in the depth maps of the novel views, as monocular depth estimators~\citep{depthanything,marigold} fail to maintain a consistent scale across viewpoints. Multiview diffusion models~\citep{muller2024multidiff,cat3d} attempt to address this by generating all novel views simultaneously for improved consistency but are limited by the number of views they can produce and demand extensive datasets and computational resources for training. In short, achieving multiview consistency in the generated images of a scene from a single-view input remains an unresolved and significant challenge.

In this paper, we propose VistaDream, a framework for 3D scene reconstruction from single-view images without the requirement of fine-tuning diffusion models. Given the single-view images as inputs, VistaDream reconstructs the scene of the given single-view image as a set of 3D Gaussian kernels~\citep{gs}, which enables us to render arbitrary novel-view images in the scene by the splatting technique. VistaDream is built upon the existing image diffusion models~\citep{fooocus,lcm} and maintains the multiview consistency of generated images by a two-stage pipeline as follows.

In the first stage, VistaDream begins by constructing a coarse 3D scaffold, achieved by zooming out the camera from the input view while applying inpainting and depth estimation. This process establishes a rough yet valuable global geometry constraint for the 3D reconstruction. By zooming out, we generate expanded views and utilize the Fooocus model~\citep{fooocus} to inpaint the black borders created by the zooming out. We further enhance this step by leveraging detailed text descriptions provided by a Visual-Language Model~\citep{llava}, which helps produce high-quality, well-defined zoomed-out images. Next, we estimate a depth map on the zoomed-out image, providing a coarse 3D geometry of the entire scene to serve as a constraint for subsequent generation. Building on this global scaffold, we then apply a warp-and-inpaint approach~\citep{genwarp,wonderjourney,wonderworld,realmdreamer} to fill gaps in the 3D scene. This step produces a rough 3D reconstruction with some inconsistencies among the generated views.

In the second stage, we introduce a novel Multiview Consistency Sampling (MCS) algorithm to resample multiview-consistent images from a pre-trained diffusion model~\citep{lcm} to refine the reconstructed 3D scene. 
In contrast to SDS~\cite{poole2022dreamfusion} which only considers one view in the regeneration process for refinement, our MCS simultaneously utilizes multiple rendered images to explicitly enforce the consistency among all images, which greatly improves the ability to model fine details, avoids averaging issues, and leads to stable convergence.
This is formulated as a constrained sampling process, where multiview consistency is enforced during the reverse diffusion process. 
We begin by rendering multiple views from the current 3D scene and introducing noise to these renderings. The MCS algorithm then denoises these images to regenerate multiview-consistent outputs. At each denoising step, we utilize the predicted $x_0$ to train a new 3DGS representation, replacing the predicted $x_0$ with a corrected $\hat{x}_0$ rendered from this 3DGS representation to enhance consistency for denoising. Our results demonstrate that this consistency rectification significantly improves the multiview consistency of the generated images, leading to higher-quality 3D scene reconstructions.


We conduct experiments on single-view images in both indoor and outdoor datasets of diverse styles. 
The results demonstrate that VistaDream, requiring no training or fine-tuning, surpasses state-of-the-art scene generation methods both qualitatively and quantitatively. Comprehensive ablation studies also validate the effectiveness of our global scaffold initialization and Multiview Consistency Sampling in enhancing scene consistency and quality.

\section{Related work}

\begin{figure*}
\begin{center}
\includegraphics[width=\linewidth]{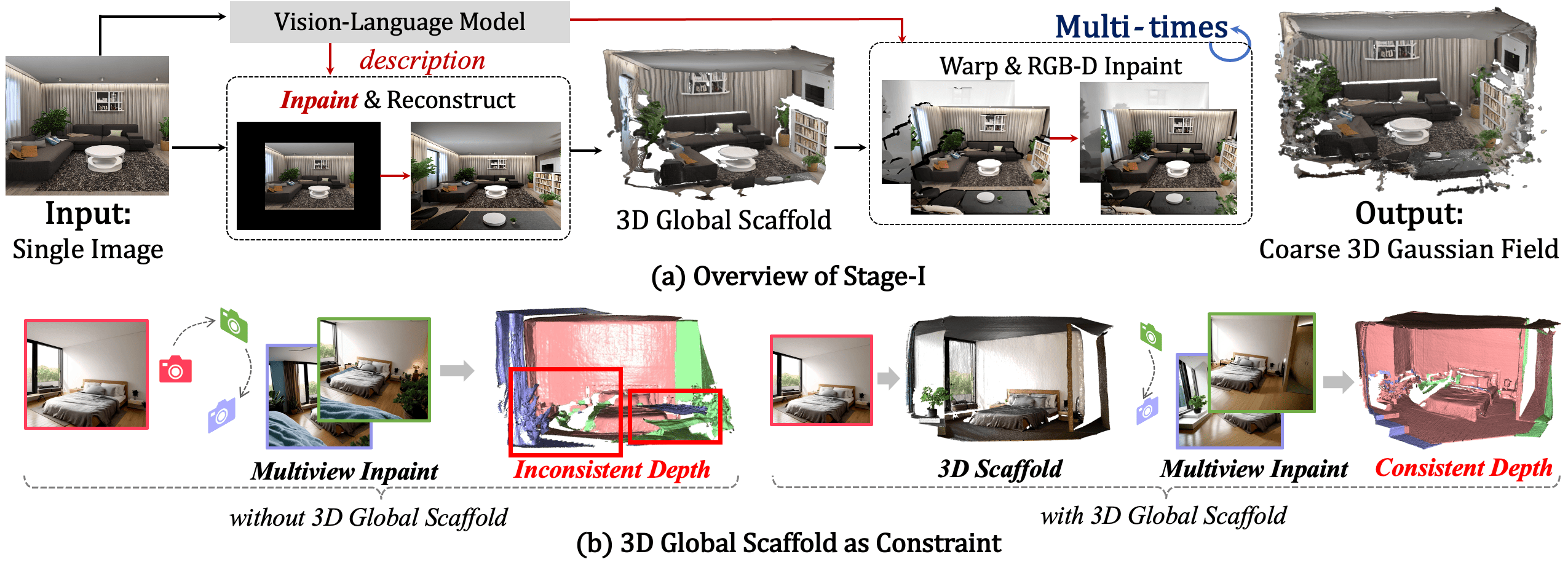}
\end{center}
\vspace{-10pt}
\caption{ 
\textit{StageI: Coarse Gaussian field reconstruction.}
(a) Given an image, VistaDream initializes a 3D global scaffold by enlarging FoV and inpainting, then iteratively inpaints the warped RGB-D images to complete a coarse Gaussian field. (b) Without a scaffold, existing models struggle to accurately connect the inpainting regions with the global scene, leading to distortion. A global scaffold provides a reliable constraint across different viewpoints, yielding correct connections between the inpainted areas and scaffold.
}
\vspace{-15pt}
\label{fig:stage1}
\end{figure*}

\textbf{Diffusion Models}.
Diffusion models have recently shown remarkable generation capabilities~\citep{ddim,ddpm}. These models gradually corrupt data into noise via a predefined Markov chain in the forward process and learn to reverse this process by progressive denoising, mapping noise distributions to data distributions. This enables effective novel data sampling or generation. Models such as Stable Diffusion~\citep{sd,sd3,sdxl,dalle3} leverage this framework for remarkable text-based image generation by scaling the model size and training data. Recent advancements fine-tune these models for tasks like depth estimation~\citep{marigold,geowizard} and image inpainting~\citep{smartbrush,genwarp,fooocus}, achieving impressive performances. Additionally, methods like Latent Consistency Models~\citep{lcm,sauer2023adversarial,yin2024one} distill pre-trained diffusion models in the latent space to enable faster one- or few-step inference. Our method is built upon existing text-to-image diffusion models~\cite{lcm,fooocus}.

\textbf{Large Vision-Language Model}.
In contrast to task-specific visual models, such as those used for segmentation~\citep{oneformer} or image captioning~\citep{clip,li2022blip,blip2}, Large Vision-Language Models (VLMs)~\citep{llava,gpt4} align visual embeddings with the latent space of Large Language Models (LLMs)~\citep{touvron2023llama,chatgpt}. By leveraging the strong knowledge priors of LLMs, VLMs enable advanced image understanding~\citep{zhang2024vision}, supporting tasks like visual question answering, image descriptions, and task decomposition. In our pipeline, we adopt the LLaVA~\cite{llava} to generate captions. Some existing works~\cite{koo2024posterior,kim2025dreamsampler} also pose constraints on the predicted $x_0$ in diffusion models for editing while our work focuses on scene generation.

\textbf{Single-view reconstruction}.
Single-view reconstruction aims to generate a 3D distribution from a single image for novel view rendering~\citep{liu2023syncdreamer,realmdreamer}. Some approaches learn to convert monocular photos into 3D objects via end-to-end training, producing 3D representations like multi-view consistent images~\citep{liu2023syncdreamer,long2024wonder3d,wang2023imagedream,li2024era3d}, meshes~\citep{liu2024one2345mesh}, neural fields~\citep{zou2024gaussain,tang2024lgm}, and tri-planes~\citep{hong2023lrm_triplane}. Alternatively, DreamFusion~\citep{poole2022dreamfusion} proposes Score Distillation Sampling (SDS) that iteratively optimizes 3D scenes through single-step sampling from noisy images rendered at various viewpoints. SDS or its variants~\citep{wang2024prolificdreamer,liang2024luciddreamer,dreamlcm} achieve lightweight 3D object generation from pre-trained 2D diffusion models. However, the randomness in SDS denoise can introduce inconsistencies among iterations, leading to averaged results~\citep{dreamlcm}.

Training end-to-end reconstruction models remains challenging for scene-level distributions due to their complexity, yielding limited fields of view and diversity~\citep{sargent2023zeronvs,cat3d}. Recent methods use inpainting models~\citep{lei2023rgbd2,chung2023luciddreamer,realmdreamer,zhang2024text2nerf} or video generation models~\citep{wang2024motionctrl,nvssolver} to iteratively complete missing regions to expand the scene scope while suffering from instability, noise, and distortion. SDS is then introduced for optimization at the cost of blurriness~\citep{realmdreamer}.
VistaDream addresses these limitations by leveraging large Vision-Language Models to enhance the reliability and diversity of scene expansion. Additionally, we propose Multi-view Consistent Sampling (MCS) to generate high-quality, consistent multi-view images directly from pre-trained diffusion models, significantly improving scene quality.

\section{Method}

\begin{figure*}
\begin{center}
\includegraphics[width=0.9\linewidth]{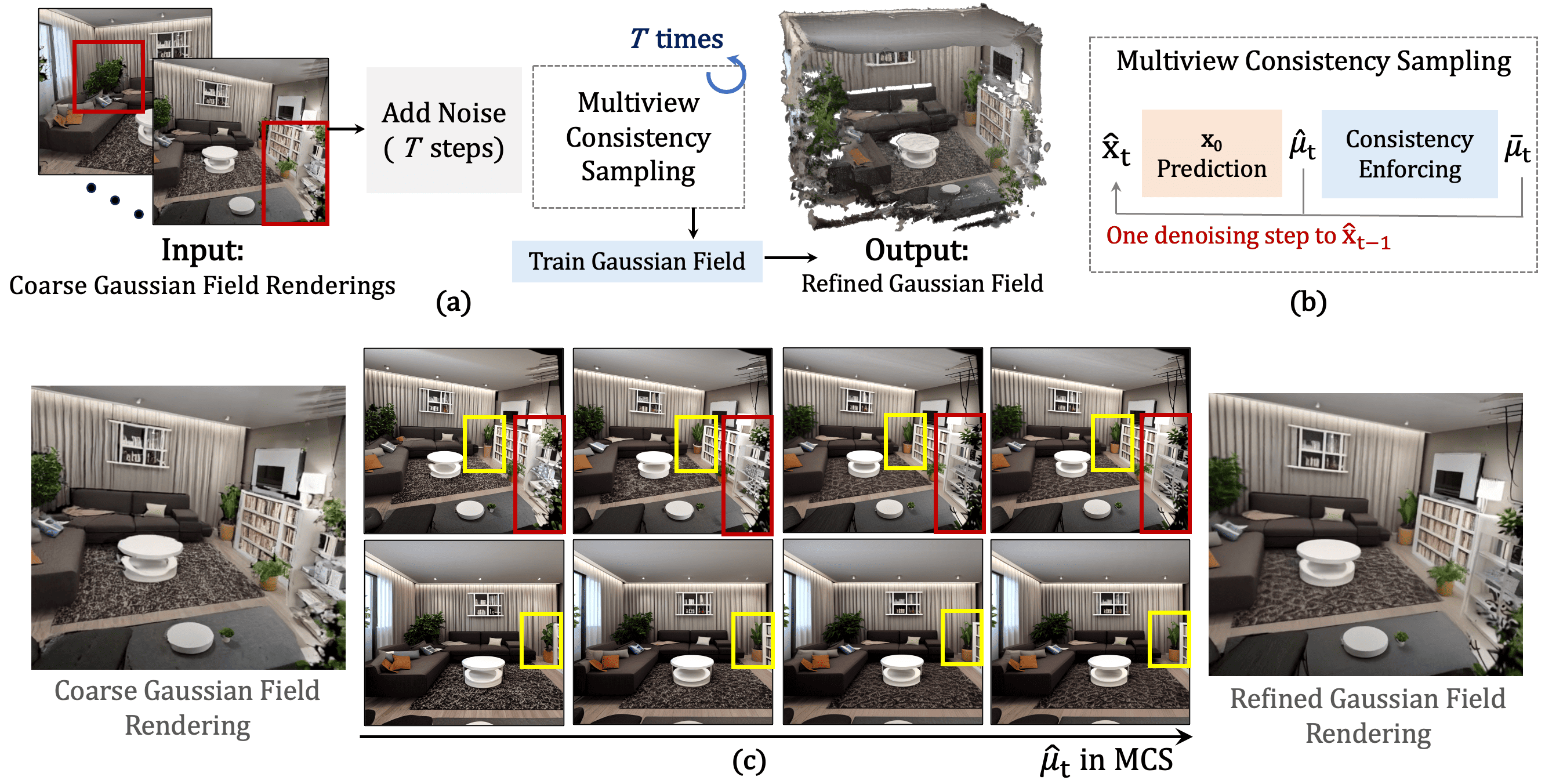}
\end{center}
\vspace{-18pt}
\caption{ 
\textit{Multiview Consistency Sampling for Scene Refinement.}
(a) We optimize the Gaussian field using high-quality, multi-view images regenerated by diffusion models. (2) The key component is the MCS algorithm, which enforces consistency during multi-view optimization. (3) A real case demonstrates that the MCS optimization process can progressively enhance the quality (red box) and consistency (yellow box) of multi-view images. Utilizing multiview images from MCS to optimize the Gaussian field can significantly enhance its quality.
}
\vspace{-10pt}
\label{fig:stage2}
\end{figure*}

Given a single-view image of a scene, the target of VistaDream is to reconstruct the 3D Gaussian field of the scene and enable novel view synthesis in the scene. VistaDream achieves this with a two-stage pipeline. The first stage builds a coarse 3D Gaussian field while the second stage refines the 3D Gaussian field with Multiview Consistency Sampling of a diffusion model.

\subsection{Coarse Gaussian field reconstruction}
In this stage, our target is to build a coarse Gaussian field from the single-view input image. In contrast to existing 3D scene generation methods~\citep{realmdreamer,genwarp} that directly apply warp-and-inpaint scheme,  as shown in Fig.~\ref{fig:stage1} (a), our method first builds a global 3D scaffold by zooming out the input view with inpainting and estimating the depth map on the zoomed-out image. Then, we apply the warp-and-inpaint scheme built on the global 3D scaffold to generate novel-view images and depth maps. Finally, we reconstruct a 3D Gaussian field from generated images and depth maps.

\textbf{Motivation}. An obvious problem in the previous warp-and-inpaint scheme is that it has difficulty in maintaining the loop consistency when the generation trajectory revisits the previously generated regions as shown in Fig.~\ref{fig:stage1} (c). In contrast, we first reconstruct a reasonable global 3D scaffold by a zoomed-out image and its estimated depth map. This 3D global scaffold constrains overall appearances and geometry for most regions, which effectively prevents the warp-and-inpainting scheme from deviating largely from the 3D scaffold to improve the multiview consistency.

\begin{figure}
\centering
\includegraphics[width=0.45\textwidth]{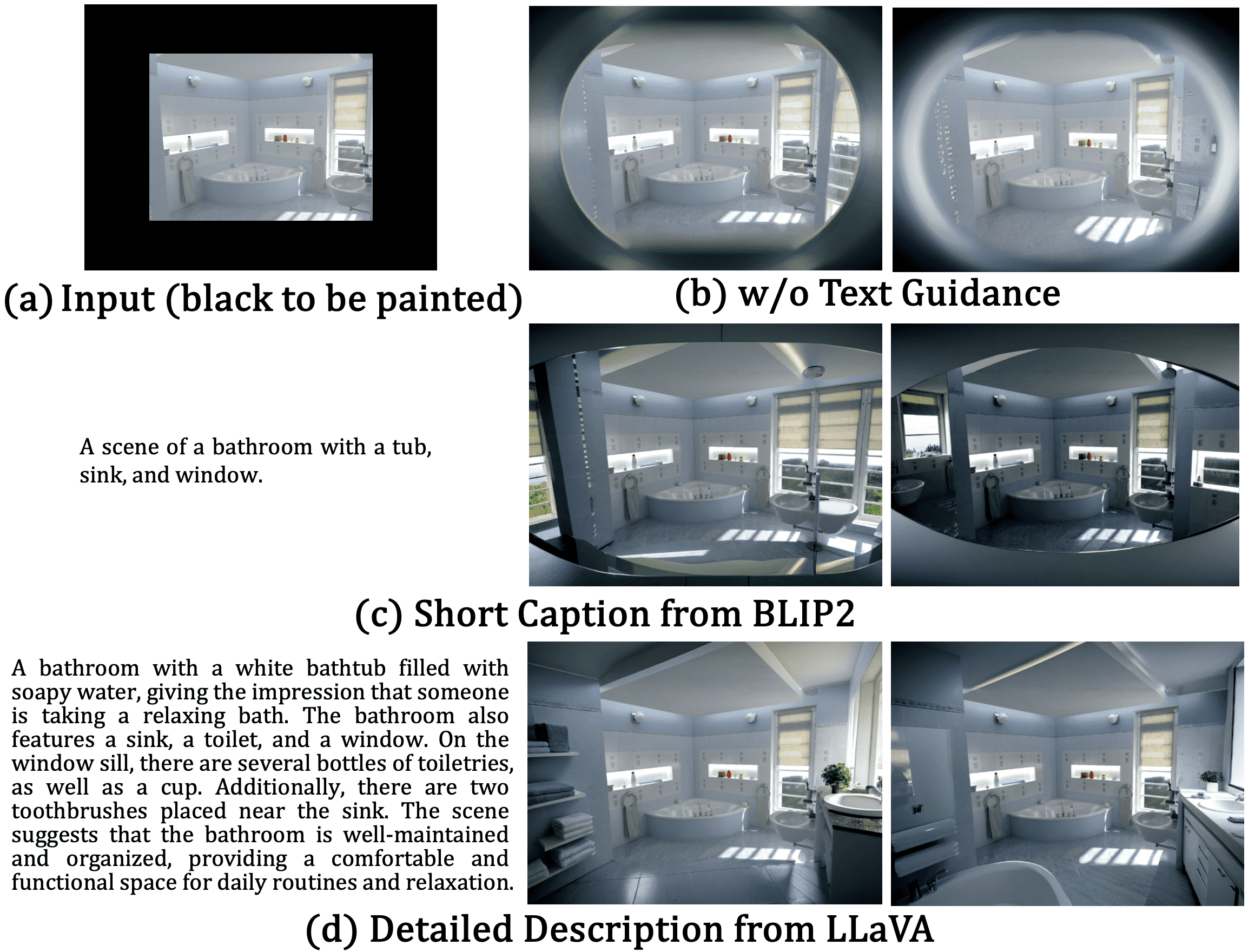}
\vspace{-5pt}
\caption{ \textit{Detailed description is vital for inpainting.} Compared to (b) empty descriptions or (c) short captions, (d) descriptions from large Vision-Language Models are more detailed, significantly enhancing the reliability of inpainting.}
\vspace{-15pt}
\label{fig:llava}
\end{figure}

\textbf{Building a 3D scaffold}.
To build the 3D scaffold, we first zoom out from the input view by changing the Field of View (FoV) of the camera as shown in Fig.~\ref{fig:stage1} (a), which leaves a large regular region for the inpainting model Fooocus~\citep{fooocus} to fill the coherent new contents on these unseen regions. 
The inpainting models usually require a text prompt to generate new content. We find that the text prompts are vital for the correct inpainting espacially for a large missing region. 
The short and simple descriptions from BLIP~\citep{blip2} leads to incorrect and low-quality new content while the detailed descriptions from LLaVA~\citep{llava} will greatly improve the inpainting quality, as shown in Fig.~\ref{fig:llava}. After generating the zoomed-out images, we apply a depth estimator~\citep{metric3dv2,depthpro} to estimate the metric depth values for all pixels on the zoomed-out image. The zoomed-out images along with the estimated depth map provide a 3D scaffold for the entire 3D scene.

\textbf{Warp-and-inpaint}. Then, based on the 3D scaffold, we apply the warp-and-inpaint scheme to generate a set of novel-view images. For a specific new viewpoint, we warp the global zoomed-out image with its depth map to this viewpoint and then fill the empty regions with the Fooocus~\citep{fooocus} model. After that, we estimate the depth map on this new viewpoint with a depth estimator~\citep{geowizard,depthpro}. 
The estimated depth map is further optimized to align with the global depth map~\citep{chung2023luciddreamer,wonderworld}. 
We repeat this process for several new viewpoints to generate a set of RGBD images. Finally, we train a 3D Gaussian field using these generated RGBD images and the global 3D scaffold as our coarse 3D Gaussian field.

\subsection{Multiview Consistency Sampling Refinement}
In the above stage, we enforce the consistency between the 3D scaffold and the novel-view images generated by the warp-and-inpaint scheme. However, there still remains inconsistency among all the generated images causing the resulting 3D Gaussian to be noisy. In this Stage II, we will improve the reconstructed 3D Gaussian field by a Multiview Consistency Sampling algorithm. Specifically, as shown in Fig.~\ref{fig:stage2} (a), we first render $N$ images on $N$ predefined viewpoints from the coarse Gaussian field, denoted by $\mathbf{x}^{(1:N)}:=\{\mathbf{x}^{(n)}|=1,...,N\}$. Then, we use the diffusion models to refine all these renderings into new images $\mathbf{\hat{x}}^{(1:N)}$ which have enhanced quality and multiview consistency. Finally, we refine the 3D Gaussian field by training on these refined renderings. The key problem here is how to refine these renderings using a diffusion model while maintaining multiview consistency because simply regenerating these images leads to inconsistent results. We address this problem with a Multiview Consistency Sampling (MCS) algorithm.

\subsubsection{Multiview Consistency Sampling}

Given the renderings $\mathbf{x}^{(1:N)}$, we first follow the forward process of the diffusion model to add noises to these renderings to get a set of noisy renderings $\mathbf{\hat{x}}^{(1:N)}_T$ where $T$ is a predefined time step. Then, to regenerate these images, we sample the Markov Chain of the reverse process $\prod_n^N \prod_t^T p_\theta(\hat{\mathbf{x}}^{(n)}_{t-1}|\hat{\mathbf{x}}^{(n)}_{t})$ while enforcing the multiview consistency between all the images $\hat{\mathbf{x}}_0^{(1:N)}$. To achieve this, we enforce the multiview consistency on every timestep by training a 3D Gaussian field to rectify the denoising direction.

\textbf{$\mathbf{x}_0$-prediction of DDPM}. Specifically, recall that on one denoising step $t$ of the DDPM~\citep{ddpm} model, the predicted noise $\epsilon_\theta(\mathbf{\hat{x}}^{(n)}_t,t)$ gives an estimation of the final denoising result $\hat{\mathbf{x}}^{(n)}_0$ by
\begin{equation}
    \hat{\mu}(\mathbf{x}^{(n)}_t,t)= \frac{1}{\bar{\alpha}_t}(\mathbf{\hat{x}}^{(n)}_t-\bar{\beta}_t\epsilon_\theta(\mathbf{\hat{x}}^{(n)}_t,t)),
    \label{eq:one-step}
\end{equation}
where $\epsilon_\theta(\mathbf{\hat{x}}^{(n)}_t,t)$ denotes the predicted noises on the $n$-th rendered view on the timestep $t$, $\hat{\mu}_t^{(n)}:=\hat{\mu}(\mathbf{x}^{(n)}_t,t)$ is the estimated $\hat{\mathbf{x}}^{(n)}_0$ from the current noisy version $\hat{\mathbf{x}}^{(n)}_t$, $\bar{\alpha}_t$ and $\bar{\beta}_t$ are predefined constants. 
Thus, one denoising step can also be written in the form of $\hat{\mu}_{t}^{(n)}$ instead of $\epsilon_\theta(\mathbf{\hat{x}}^{(n)}_t,t)$ by
\begin{equation}
    \hat{\mathbf{x}}_{t-1}^{(n)} = s_t \hat{\mathbf{x}}_{t}^{(n)} + d_t \hat{\mu}_t^{(n)}+\sigma_t \mathbf{\epsilon}, \mathbf{\epsilon}\sim \mathcal{N}(0,I),
    \label{eq:mu-denoise}
\end{equation}
where $s_t$, $d_t$, $\sigma_t$ all are predefined constants and $\mathbf{\epsilon}$ is a noise sampled from the standard Gaussian distribution. 
Eq.~(\ref{eq:mu-denoise}) indicates that the denoising direction is determined by the $\mathbf{\hat{\mu}}_{t}^{(n)}$. Thus, the key idea of MCS is to rectify $\mathbf{\hat{\mu}}_t^{(1:N)}:=\{\mathbf{\hat{\mu}}_{t}^{(n)}\}$ to new $\mathbf{\tilde{\mu}}_t^{(1:N)}$ and then use the rectified $\mathbf{\tilde{\mu}}^{(1:N)}_t$ for denoising in Eq.~(\ref{eq:mu-denoise}).

\textbf{Enforcing consistency}. Since $\mathbf{\hat{\mu}}^{(1:N)}_t$ is an estimation of the noisy free $\hat{\mathbf{x}}^{(1:N)}_0$, we enforce the multiview consistency between them by training a 3D Gaussian field on the noisy-free $\mathbf{\hat{\mu}}^{(1:N)}_t$. Then, we render the images on this temporal 3D Gaussian field, which are denoted by $\bar{\mathbf{\mu}}_t^{(n)}$. Then, the rectified $\mathbf{\tilde{\mu}}^{(1:N)}_t$ are computed by
\begin{equation}
    \mathbf{\tilde{\mu}}^{(n)}_t = w_t \gamma_t^{(n)} \bar{\mathbf{\mu}}_t^{(n)} + (1-w_t) \mathbf{\hat{\mu}}_t^{(n)},
    \label{eq:w}
\end{equation}
where $\gamma_t$ stands for $std(\mathbf{\hat{\mu}}_t)/std(\bar{\mathbf{\mu}}_t)$ to avoid over-exposure~\citep{stdalign}, $w_t$ is a predefined weight to balance between the denoising results $\mathbf{\hat{\mu}}^{(1:N)}_t$ and the rendered multiview consistent $\mathbf{\bar{\mu}}^{(1:N)}_t$. $w_t$ determines how much multiview consistency is imposed on the denoising process. Learning a 3D Gaussian field forces the multiview consistency to get $\mathbf{\bar{\mu}}^{(1:N)}_t$ but may oversmooth some regions. Directly utilizing the denoising directions from $\mathbf{\hat{\mu}}^{(1:N)}_t$ produces images with more details but less multiview consistency. Therefore, we set the $w_t$ to balance the denoising directions between $\mathbf{\bar{\mu}}^{(1:N)}_t$ and $\mathbf{\hat{\mu}}^{(1:N)}_t$. We repeat this process for every denoising step to get the refined renderings $\hat{\mathbf{x}}_0^{(1:N)}$. Then, these refined renderings are used for the refinement of the coarse 3D Gaussian field to improve the rendering quality as shown in Fig.~\ref{fig:stage1} (c).

\textbf{Discussion}. Previous methods~\citep{realmdreamer,genwarp,wonderjourney} mainly focus on enforcing the consistency between the input image and a single generated image in a sequential manner, which struggles to maintain consistency on a long trajectory. Our MCS allows the simultaneous generation of multiple novel-view images and enforcement of consistency among all generated images, which does not suffer from consistency lost in sequential modeling. Thus, MCS generates more high-quality and consistent images than baseline methods and improves the quality of single-view 3D reconstruction. 
An alternative way is to adopt the SDS-based refinement method~\cite{poole2022dreamfusion,realmdreamer}. However, the SDS method only considers one rendered view for one denoising step in the optimization, which often tends to average the generated contents. In comparison, our MCS simultaneously considers multiple rendered views to maintain multiview consistency and thus improves the consistency and generation quality.

\begin{figure*}
\begin{center}
\includegraphics[width=0.85\linewidth]{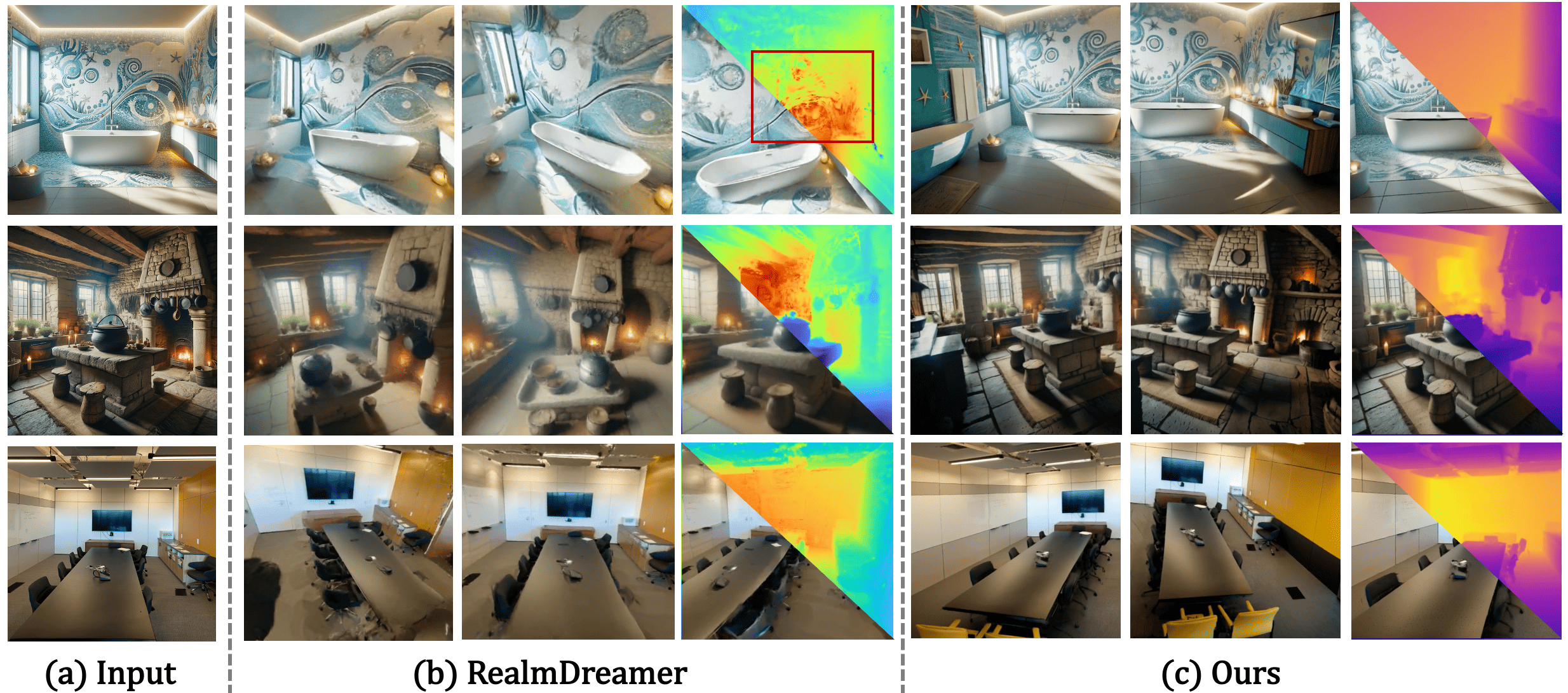}
\end{center}
\vspace{-15pt}
\caption{ 
\textit{Qualitative comparisons between RealmDreamer~\cite{realmdreamer} and our method.}
Given (a) a single input image, both (b) RealmDreamer and (c) VistaDream (Ours) reconstruct the corresponding 3D Gaussian Field through a coarse-to-fine strategy. In the third column of each method, we visualize a mixture of rendered images and depth maps of the scene.}
\label{fig:realm}
\vspace{-10pt}
\end{figure*}

\begin{figure*}
\begin{center}
\includegraphics[width=0.85\linewidth]{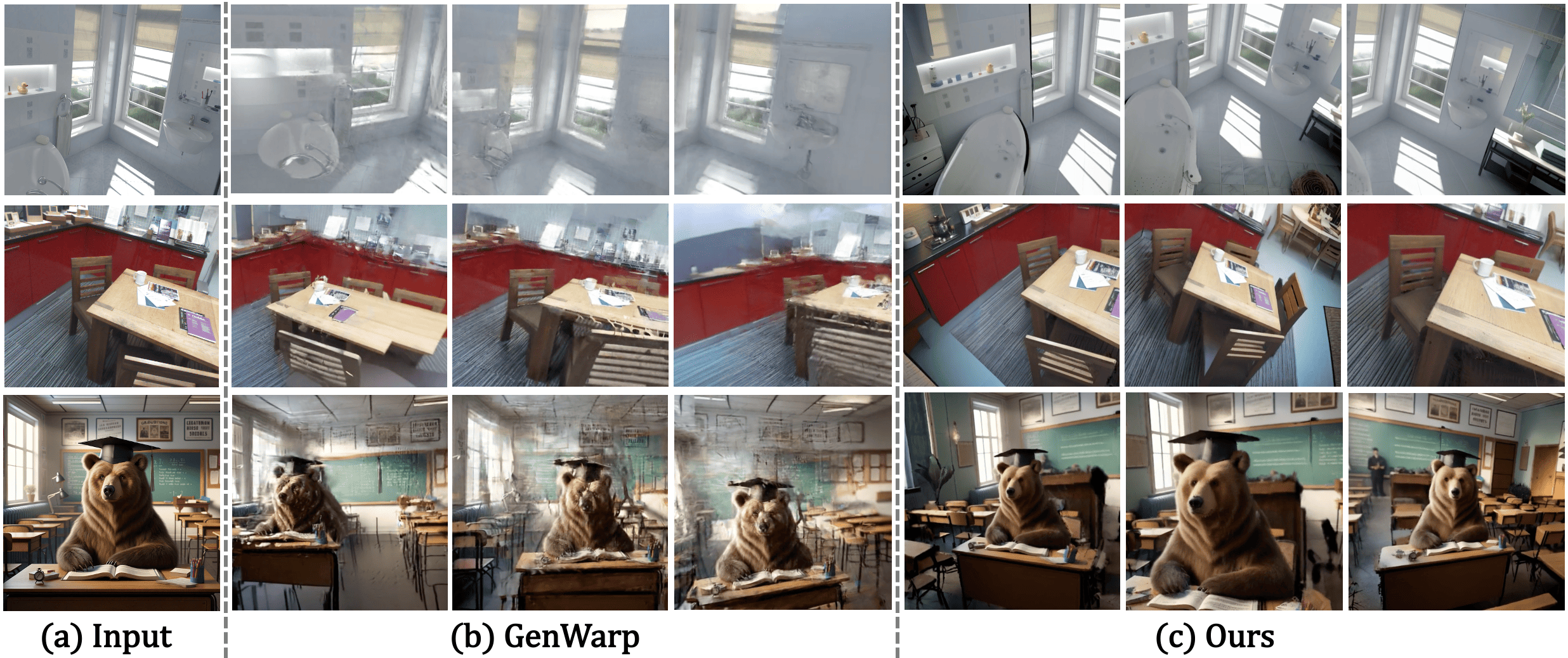}
\end{center}
\vspace{-15pt}
\caption{ 
\textit{Qualitative comparisons between GenWarp~\cite{genwarp} and our method.}
Given (a) a single input image, GenWarp generates a Gaussian field by applying InstantSplat~\cite{instantsplat} to the estimated multiview images. We present the rendered novel views from the scenes reconstructed by (b) GenWarp and (c) VistaDream (Ours).}
\vspace{-10pt}
\label{fig:genwarp}
\end{figure*}

\begin{figure*}
\begin{center}
\includegraphics[width=0.85\linewidth]{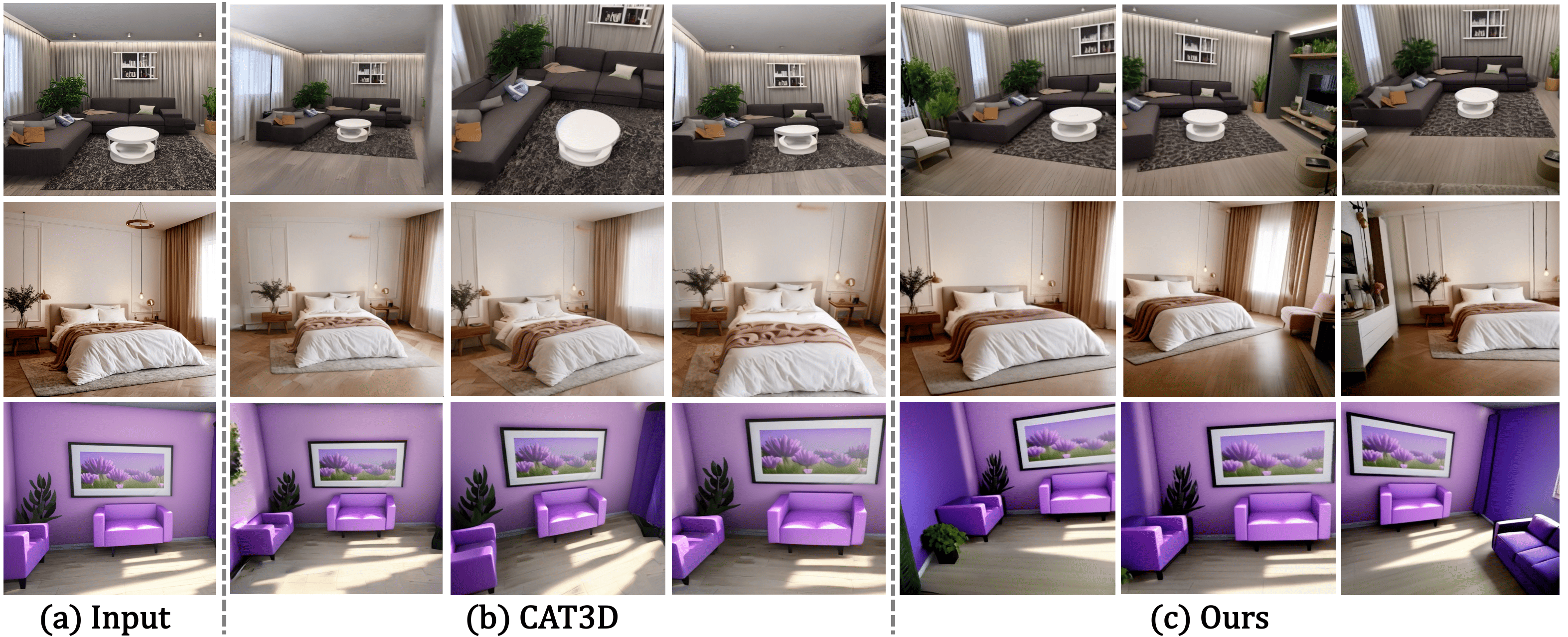} 
\end{center}
\vspace{-15pt}
\caption{ 
\textit{Qualitative comparisons between CAT3D~\cite{cat3d} and our method.}
Given (a) a single input image, (b) CAT3D conducts multi-view diffusion by conditioning on the input image and reconstructs the scene with these images by Zip-NeRF~\cite{barron2023zip}. (c) VistaDream (Ours) develops a two-stage framework for single-view scene reconstruction, achieving larger scenes with more stuffs.}
\label{fig:cat3d}
\vspace{-10pt}
\end{figure*}

\begin{table*}
\begin{center}
\resizebox{0.75\linewidth}{!}{
\begin{tabular}{lcccccc}
Method       & \textit{\textbf{Train}}                      & \textbf{Noise-Free↑} & \textbf{Edge↑} & \textbf{Sturcture↑} & \textbf{Detail↑} & \textbf{Quality↑} \\ \hline\hline
GenWarp~\cite{genwarp}      & \checkmark                                            & 0.496                & 0.062          & 0.333               & 0.445            & 0.338             \\
RealmDreamer~\cite{realmdreamer} & \checkmark                                            & 0.847                & 0.129          & 0.325               & 0.835            & 0.431             \\
CAT3D~\cite{cat3d}        & \checkmark                                            & \textbf{0.962}       & 0.253          & 0.464               & \textbf{0.976}   & \textbf{0.765}    \\
\rowcolor[gray]{0.9} 
Ours-Coarse  &  & 0.909                & \underline{0.285}    & \underline{0.542}         & \underline{0.967}      & 0.709             \\
\rowcolor[gray]{0.9} 
Ours         &  & \underline{0.951}          & \textbf{0.342} & \textbf{0.611}      & 0.951            & \underline{0.733}       \\
\end{tabular}}
\vspace{-5pt}
\caption{\textit{Quantitative evaluations} on renderings from the reconstructed scenes.}
\vspace{-15pt}
\label{tab:main}
\end{center}
\end{table*}

\begin{figure*}
\begin{center}
\includegraphics[width=1.\linewidth]{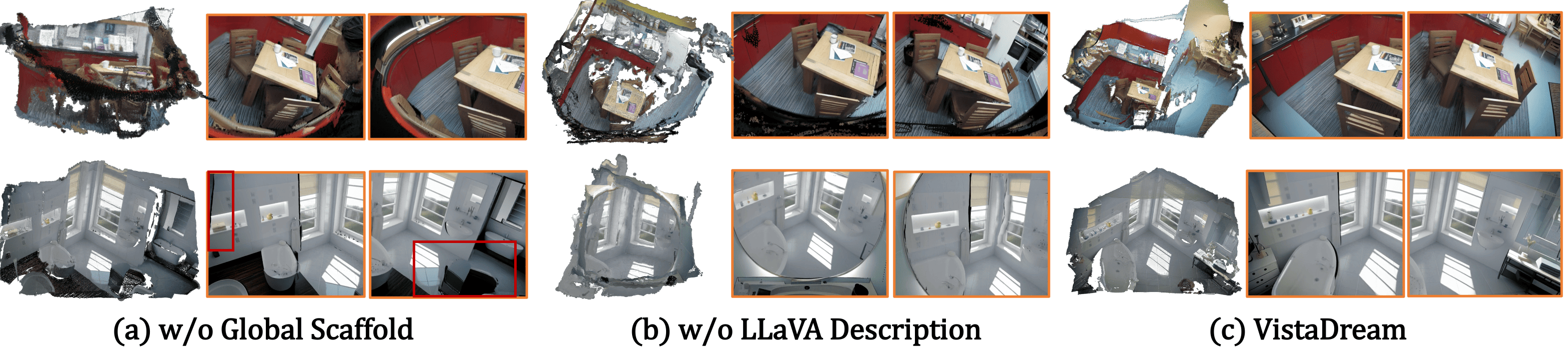}
\end{center}
\vspace{-15pt}
\caption{ 
\textit{Ablating 3D global scaffold in coarse scene reconstruction}
(a) Without the 3D global scaffold, the reconstructed scene shows distortions and generates unwanted human regions.
(b) Reconstructing the coarse scene with the guidance of short captions from BLIP2 yields telescope-like or mirror-like images. (c) Using LLaVA for description greatly improves the generated quality.
}
\vspace{-10pt}
\label{fig:blip}
\end{figure*}

\begin{figure*}
\begin{center}
\includegraphics[width=0.93\linewidth]{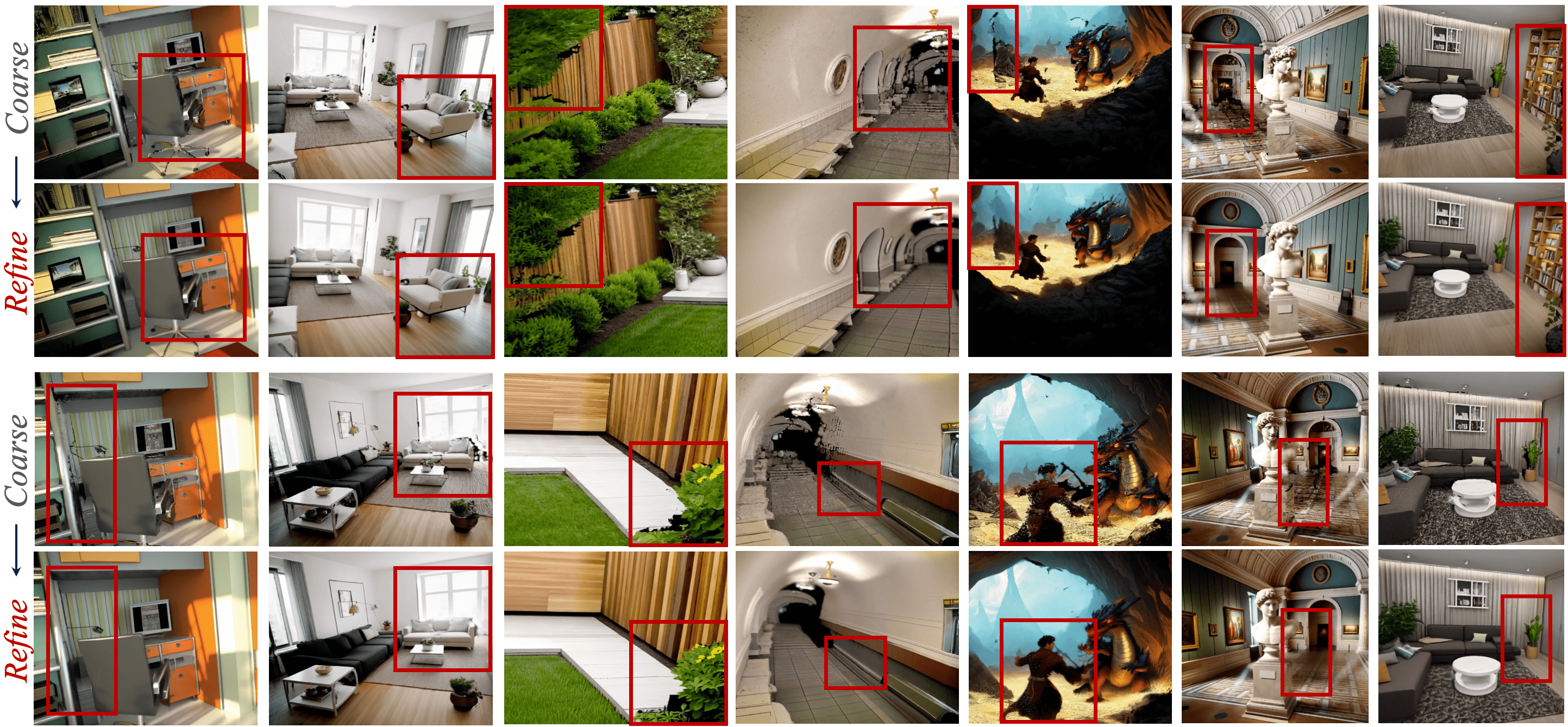}
\end{center}
\vspace{-12pt}
\caption{ 
\textit{Effectiveness of MCS refinement.}
(a) The coarse Gaussian field contains some noisy and messy objects as marked by red boxes. (b) After our MCS refinement, the rendering results demonstrate improved quality.}
\vspace{-5pt}
\label{fig:coarse}
\end{figure*}

\begin{figure*}
\begin{center}
\includegraphics[width=.93\linewidth]{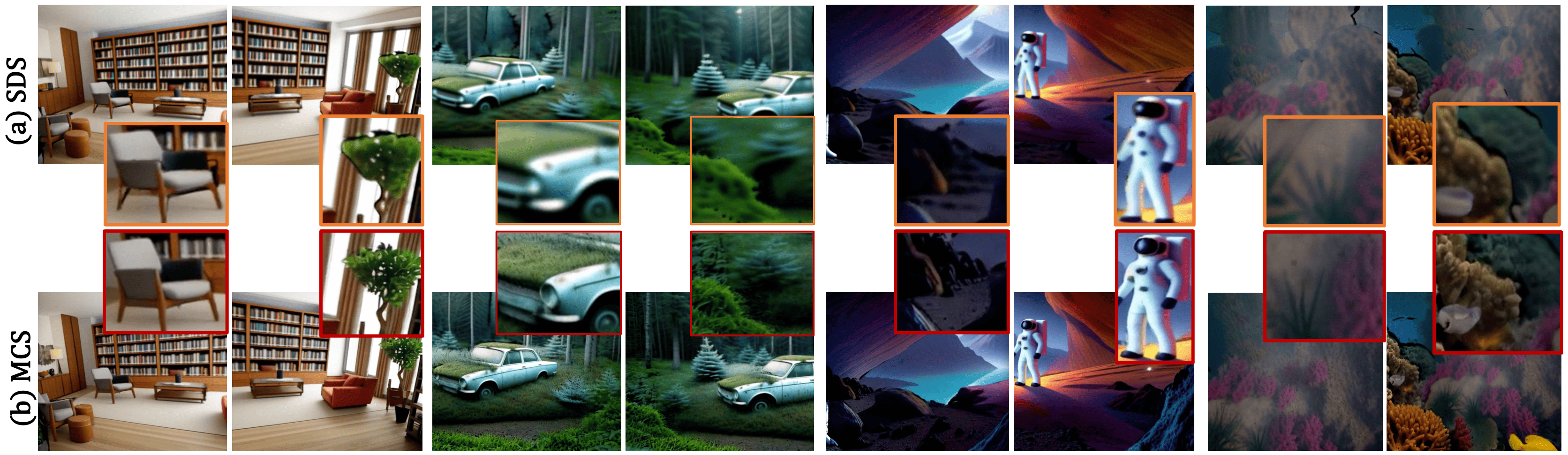}
\end{center}
\vspace{-12pt}
\caption{ 
\textit{Scene refinement via Score Distillation Sampling (SDS) or our Multiview Consistency Sampling (MCS)}.
(a) The SDS refinement yields an overly smooth 3D scene, leading to blurry and inconsistent artifacts in the rendered images. (b) In contrast, our method produces enhanced qualities and better realism.
}
\vspace{-14pt}
\label{fig:sds}
\end{figure*}

\section{Experiments}
\subsection{Experimental protocol}
\textbf{Datasets.} 
We adopt 34 single-view images from baseline methods~\citep{realmdreamer,nvssolver}, copyright-free online photos, and generated models~\citep{sd3} for evaluation. The images cover both indoor and outdoor, real and simulated scenes. Among these, 11 images from RealmDreamer~\citep{realmdreamer} were used in the quantitative comparisons. The RGB and depth videos rendered from these scenes using our method are provided in the supplementary material.

\textbf{Baselines.} We adopt RealDreamer~\citep{realmdreamer}, GenWarp~\citep{genwarp}, and CAT3D~\citep{cat3d} as the baseline methods. Given an input image,  RealDreamer~\citep{realmdreamer} trains inpainting networks for both RGB and depth images to iteratively extend a Gaussian field as the scene representation, which is then refined through an optimization process with a Score Distillation Sampling (SDS)~\citep{sds} loss.
GenWarp~\citep{genwarp} trains a network to generate novel-view images of the same scene and leverages InstantSplat~\citep{instantsplat} to reconstruct the scene with a Gaussian field.
CAT3D~\citep{cat3d} trains a multi-view diffusion model to simultaneously sample the multi-view images by conditioning on the input image and then reconstructing the scene with Zip-NeRF~\citep{barron2023zip}. For GenWarp~\citep{genwarp}, we adopt their official implementation to test on these scenes. For CAT3D~\citep{cat3d} and RealmDreamer~\citep{realmdreamer}, since they have not released the codes yet, we use the results provided in their project pages.

\textbf{Metrics.} Inspired by CLIP-IQA~\citep{clipiqa} and WonderJourney~\citep{wonderjourney}, we employ VLM, specifically LLaVA~\citep{llava}, to evaluate the quality of the multiview images rendered from the reconstructed scenes on five aspects: noise level, edge clarity, structure, detail, and overall quality. Details of the LLaVA-IQA metric are given in Sec.~\ref{sec:llavaiqa} of the appendix.

\textbf{Implementation details.} We conducted all experiments on a single 4090 GPU (24G). It takes $5\sim8$ minutes to reconstruct a single scene. More implementation details are included in Sec.~\ref{sec:impldetails} of the appendix.

\subsection{Comparisons with baselines}
In Fig.~\ref{fig:realm}-\ref{fig:cat3d}, we show qualitative results of the proposed method and the baseline methods RealDreamer~\citep{realmdreamer}, GenWarp~\citep{genwarp}, and CAT3D~\citep{cat3d} across various scenes. More image/text-to-scene results and interactive demos are provided in the supplementary material. The quantitative results in Table~\ref{tab:main} show that our VistaDream without any finetuning on the single-view scene reconstruction task demonstrates significant improvements over RealDreamer and GenWarp and achieves comparable qualities as CAT3D which has been extensively trained on the single-view scene reconstruction task. 

In Fig.~\ref{fig:realm}, RealmDreamer exhibits significant distortion and noise due to the inconsistency introduced by warp-and-inpaint and shows blurry renderings due to the SDS refinement.
In Fig.~\ref{fig:genwarp}, the multi-view images generated by GenWarp exhibit noticeable inconsistencies, leading to noise and distortion in the reconstructed scenes.
In Fig.~\ref{fig:cat3d}, the multi-view images generated by CAT3D exhibit high quality and strong consistency by training on a large-scale multiview dataset, enabling the reconstruction of reliable and clear scenes. 
In comparison to existing methods, VistaDream enlarges the initial scene scope with VLM-assisted inpainting, which improves the consistency and stability of subsequent inpainting. Furthermore, we ensured both multi-view consistency and quality enhancement during scene optimization, ultimately yielding accurate and realistic reconstructions.

\subsection{More analysis}
More analysis about $w_t$ setting in Eq.~\ref{eq:w} and failure case are provided in the supplementary material.

\textbf{Ablating global scaffold construction}. 
In Fig.~\ref{fig:blip}, we conduct ablation studies of the 3D global scaffold construction in the first stage. Without the 3D global scaffold provided by inpainting, the reconstructed scene may suffer from distortions caused by the unstable inpainting from any viewpoint~\citep{wonderjourney}. However, utilizing global scaffold is not straightforward; without the detailed descriptions provided by LLaVA~\citep{llava}, the inpainting model tends to produce significant distortions in the scaffold, such as large rings. The LLaVA-assisted inpainting for building scaffold significantly improved the stability and diversity of scene reconstruction.

\textbf{Effectiveness of MCS refinement}. 
As shown in Fig.~\ref{fig:coarse}, the rendering of coarse Gaussian fields exhibits noticeable noise and artifacts, including distorted object boundaries and chaotic structures in complex regions. 
After MCS refinement, the accuracy and overall coherence of the scene have improved, allowing for the rendering of high-quality novel views, though minor detail blurring may occur to enforce multi-view consistency. The quantitative results in Table~\ref{tab:main} further support these analyses.

\textbf{Compare MCS with SDS refinement}. 
Score Distillation Sampling (SDS)~\citep{sds} is a commonly used scene optimization technique that iteratively refines the single-view renderings by one-step diffusion. However, SDS only considers one input view, it tends to average the generation results for consistency, yielding blurry results~\citep{dreamlcm} as shown in Fig.~\ref{fig:sds} (a). 
The proposed Multi-view Consistency Sampling simultaneously samples multi-view images by explicitly enforcing consistency, yielding high-quality and coherent multi-view images. These images achieve accurate and realistic scene optimization as shown in Fig.~\ref{fig:sds} (b).

\section{Conclusion}
We propose VistaDream, a two-stage framework for 3D scene reconstruction from a single image. In the first stage, we enhance the stability of scene reconstruction by introduction a VLM-assisted global scaffold. In the second stage, we introduce Multi-view Consistency Sampling to sample high-quality and consistent multi-view images for scene optimization. Experimental results demonstrate that our method requires no fine-tuning on the single-view scene reconstruction task but achieves superior qualitative and quantitative results compared to the baseline methods.

{
\small
\bibliographystyle{ieeenat_fullname}
\bibliography{main}
}

\appendix
\section{Appendix}

\subsection{Implementation details of VistaDream}
\subsubsection{Coarse Gaussian field generation}
\textbf{Image description with VLM}.
\label{sec:impldetails}
We use LLaVA~\citep{llava} to generate a detailed description for the input image. The LLaVA prompt is set as: ``\textit{$\langle$image$\rangle$ USER: Detaily imagine and describe the scene this image is taken from? ASSISTANT: This image is taken from a scene of}''. The continuation of the LLaVA response is used as the image description and fed to inpainting models in the Coarse scene reconstruction.

\textbf{Building a 3D scaffold}. The input image is enlarged by extending in four directions and inpainted using Fooocus~\citep{fooocus} with LLaVA image description. Subsequently, we can recover the per-pixel depth $d$ and image focal length $f$ using a metric depth estimator such as Metric3Dv2~\citep{metric3dv2} or Depth-Pro~\citep{depthpro}, thereby recovering the 3D points corresponding to each pixel. We follow the default hyperparameter settings of the above models.
Afterward, we follow pixelSplat~\citep{pixelsplat} to construct Gaussian kernels for each pixel: the \textit{xyz} property of the Gaussian kernels is its 3D position, the \textit{RGB} property comes from the pixel color, the \textit{opacity} property is set to a constant, the \textit{rotation} property is an identity matrix, and the scale is set to $d/\sqrt{2}f$.
To avoid trailing artifacts, we eliminate kernels in object boundary regions based on depth variation judgment~\citep{freereg} and then optimize the remaining Gaussian kernels by 100 iterations~\citep{gs}. For Gaussian kernel optimization, we set the learning rate of the \textit{xyz} property to 3e-4, \textit{RGB} to 5e-4, \textit{scale} to 5e-3, \textit{opacity} to 5e-2, \textit{rotation} to 1e-3.

\textbf{Warp-and-inpaint}. 
After scaffold initialization, we establish a spiral camera trajectory. Then we select the viewpoint with the largest missing regions to render both the partial RGB image and depth map. The RGB image is inpainted by Fooocus~\citep{fooocus}. Taking the completed image as the condition, 
we use a model $\phi$ to estimate its depth map and optimize the depth for smoothly connecting to the existing Gaussian Field.
We have two strategies for setting $\phi$. The first strategy uses a diffusion model-based GeoWizard~\cite{geowizard} to estimate depth. To ensure smooth connections, we introduce a loss between the estimated depth and the rendered one at each denoise step~\cite{wonderjourney}. The second strategy employs a feedforward depth estimation model, DepthPro~\cite{depthpro}, to estimate image depth. We linearly align the estimated depth with the rendering one, and further optimize the estimation through residual smoothing~\cite{chung2023luciddreamer}. The first strategy is more time-consuming but yields better results, while the second strategy is faster but may introduce distortions. In different cases, we adopt the strategy that provides better visual outcomes.

Then, we construct a set of Gaussian kernels on the completed RGB-D regions as above. We filter them with two additional checks: 1) \textit{Occlusion avoidance}: We project the Gaussian center onto already processed viewpoints, and if its depth is less than the original depth at any viewpoint, it is discarded. 2) \textit{Boundary exclusion}: we remove the kernels on the object boundaries as mentioned above.
The remaining kernels are integrated into the Gaussian field. This is followed by a 256-step scene optimization process. The above ``warp-and-inpaint'' process is iteratively executed several times to obtain the coarse Gaussian field.

\subsubsection{Multiview Consistency Sampling for refinement}
\textbf{Multi-view Consistency Sampling}.
In our implementation, we uniformly sample $N=8$ views along the spiral trajectory, with an image resolution of $512\times512$.
Afterward, we encode and add $T=10$ steps of noise to each view by a 50-step DDPM sampler~\citep{ddpm}.
We use the Latent Consistency Model of Stable Diffusion (LCM-SD)~\citep{lcm} for noise prediction for its strong performance following DreamLCM~\citep{dreamlcm}. We remove Classifier Free Guidance (CFG) in LCM and find better results without it. 
We perform weighted rectification of Eq.~\ref{eq:w} on the noise map $\epsilon$ in practice, which has a linear relationship with $\mu$ according to Eq.~\ref{eq:one-step}.
In each sampling step of MCS, we use the denoising multi-view images to optimize a copy of the coarse Gaussian field by 2560 steps to enforce consistency, where we set a smaller learning rate of \textit{xyz} in Gaussian kernels, specifically 1e-4, to avoid geometry distortions.

\textbf{Gaussian field refinement}. In our implementation, we optimize the coarse Gaussian Field by 2560 steps with the refined multi-view images and enlarged input image.

To run VistaDream within a 24GB VRAM limit, we need to allocate some time for model swapping. Specifically, we transfer only the currently active model to the GPU while keeping the others in CPU memory. This ensures efficient memory usage to maintain the overall workflow's integrity.

\begin{figure*}
\begin{center}
\includegraphics[width=0.9\linewidth]{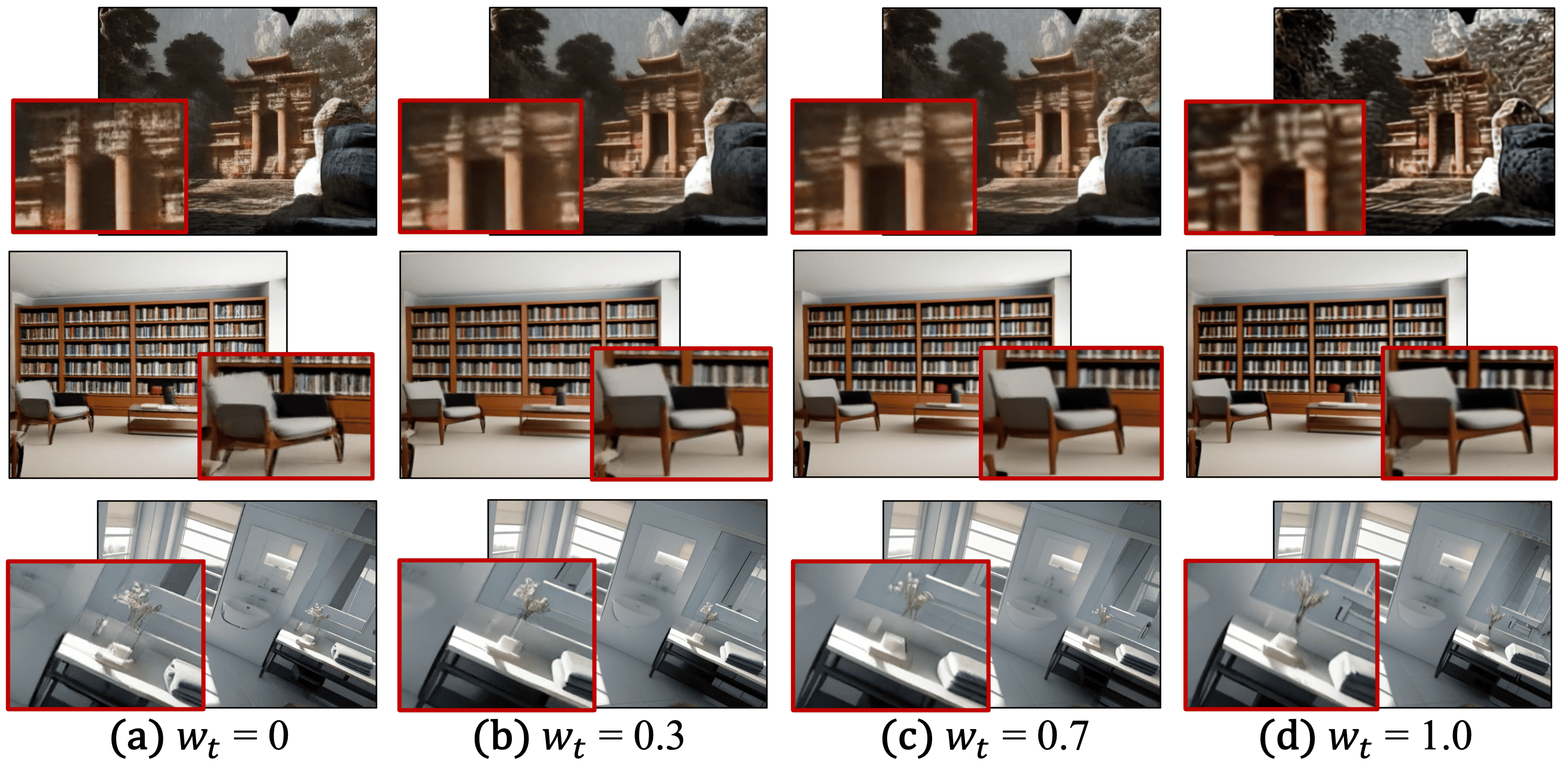}
\end{center}
\caption{ 
\textit{Set different $w_t$ in Eq.~\ref{eq:w}.}
When $w_t$ is set to 0, the optimization of the Gaussian scene lacks multi-view consistency, leading to chaotic reconstructions and noisy details. As $w_t$ increases, multi-view consistency improves, facilitating a more accurate optimization of the Gaussian field but slightly loses some details. 
}
\label{fig:w}
\end{figure*}

\subsection{More analysis}
\textbf{Choice of $w_t$ in Eq.~\ref{eq:w}}. 
In Fig.~\ref{fig:w}, we show qualitative results using different $w_t$. When $w_t=0$, the multi-view images are optimized independently to obtain high-quality but inconsistent images, yielding noisy and chaotic details after optimizing the scene. As the value of $w_t$ increases, the consistency guidance is strengthened, leading to more accurate scene optimization. However, some finer details may be lost in this process to satisfy consistency. Empirically, we found that setting $w_t$ between 0.3 and 0.8 achieves optimal results, striking a balance between detail enhancement and overall coherence.
In this section, as well as in the ``Compare MCS with SDS refinement'' section of the main text, we did not optimize the scene based on the input image, in order to more accurately reflect the effects of SDS and MCS.

\subsection{LLaVA-IQA metric details}
\label{sec:llavaiqa}
Given a set of rendered images, we perform the Image Quality Assessment using LLaVA~\citep{llava}, called LLaVA-IQA. The prompt is designed as:
``\textit{$\langle$image$\rangle$ USER: $\langle$question$\rangle$, just answer with yes or no? ASSISTANT:}''.
The $\langle$\textit{question}$\rangle$ placeholder is replaced according to different evaluation purposes as follows:
\begin{itemize}
    \item For noise level (\textbf{Noise-Free}): ``\textit{Is the image free of noise or distortion}''
    \item For edge clarity(\textbf{Edge}): ``\textit{Does the image show clear objects and sharp edges}''
    \item For scene structure(\textbf{Structure}): ``\textit{Is the overall scene coherent and realistic in terms of layout and proportions in this image}''
    \item For image details(\textbf{Detail}): ``\textit{Does this image show detailed textures and materials}''
    \item For overall image quality(\textbf{Quality}): ``\textit{Is this image overall a high-quality image with clear objects, sharp edges, nice color, good overall structure, and good visual quality}''
\end{itemize}
We then calculate the proportion of “yes” responses as the evaluation result.

We use 11 scenes from RealmDreamer~\citep{realmdreamer} for quantitative assessment, including \textit{bathroom}, \textit{bear}, \textit{bedroom}, \textit{bust}, \textit{kitchen}, \textit{living-room}, \textit{car}, \textit{lavender}, \textit{piano}, \textit{victorian}, and \textit{steampunk}. For each scene, we sample 50 viewpoints along the reconstruction trajectory for rendering and evaluation.

\subsection{Additional qualitative results}

\begin{figure*}
\begin{center}
\includegraphics[width=0.95\linewidth]{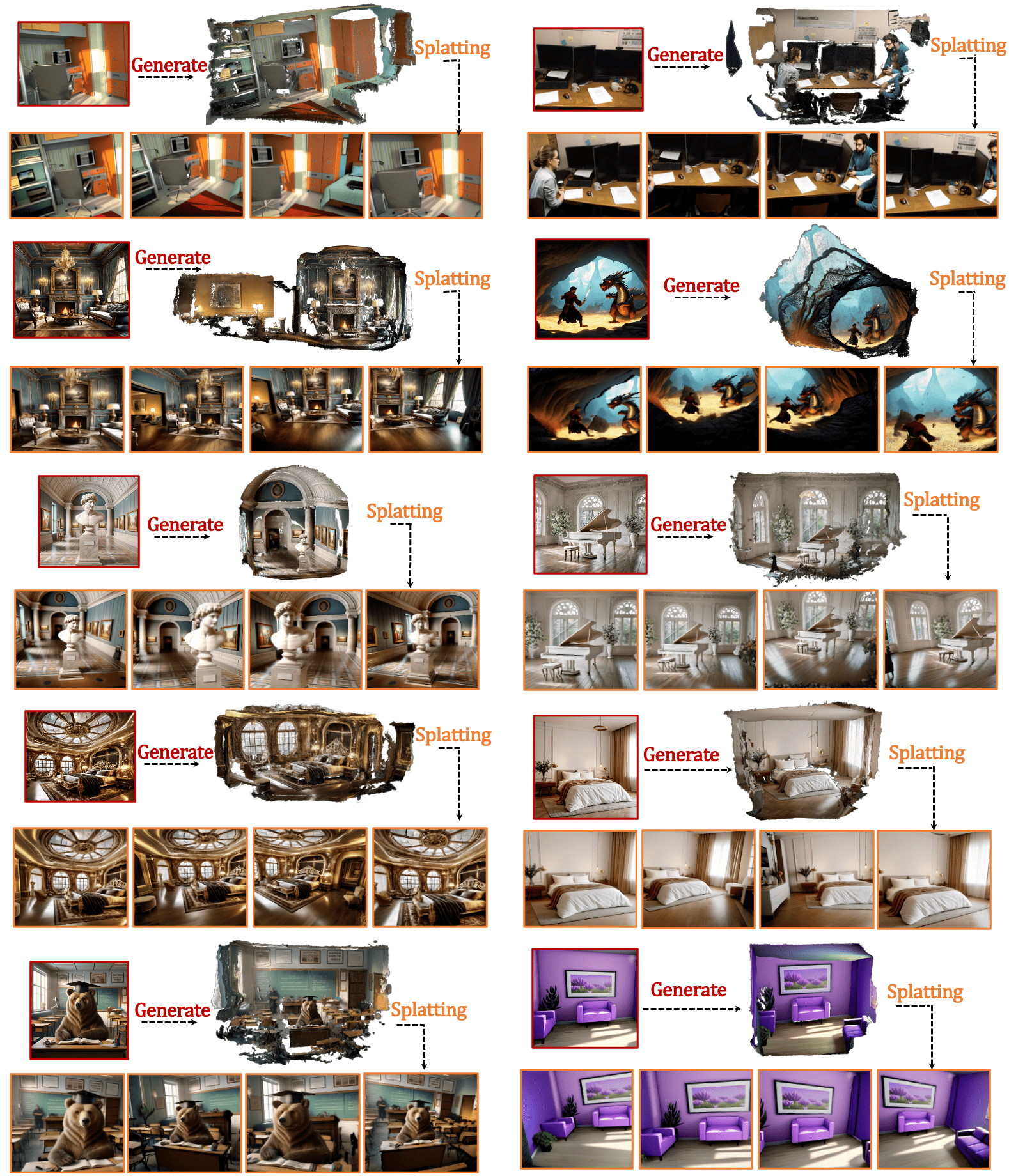}
\end{center}
\vspace{-5pt}
\caption{ 
\textit{Image-to-3D scenes.}
In each example, VistaDream generates a 3D Gaussian field based on the input image (red box), which is capable of rendering novel view images (orange box).
}
\label{fig:main}
\end{figure*}

\begin{figure*}
\begin{center}
\includegraphics[width=0.95\linewidth]{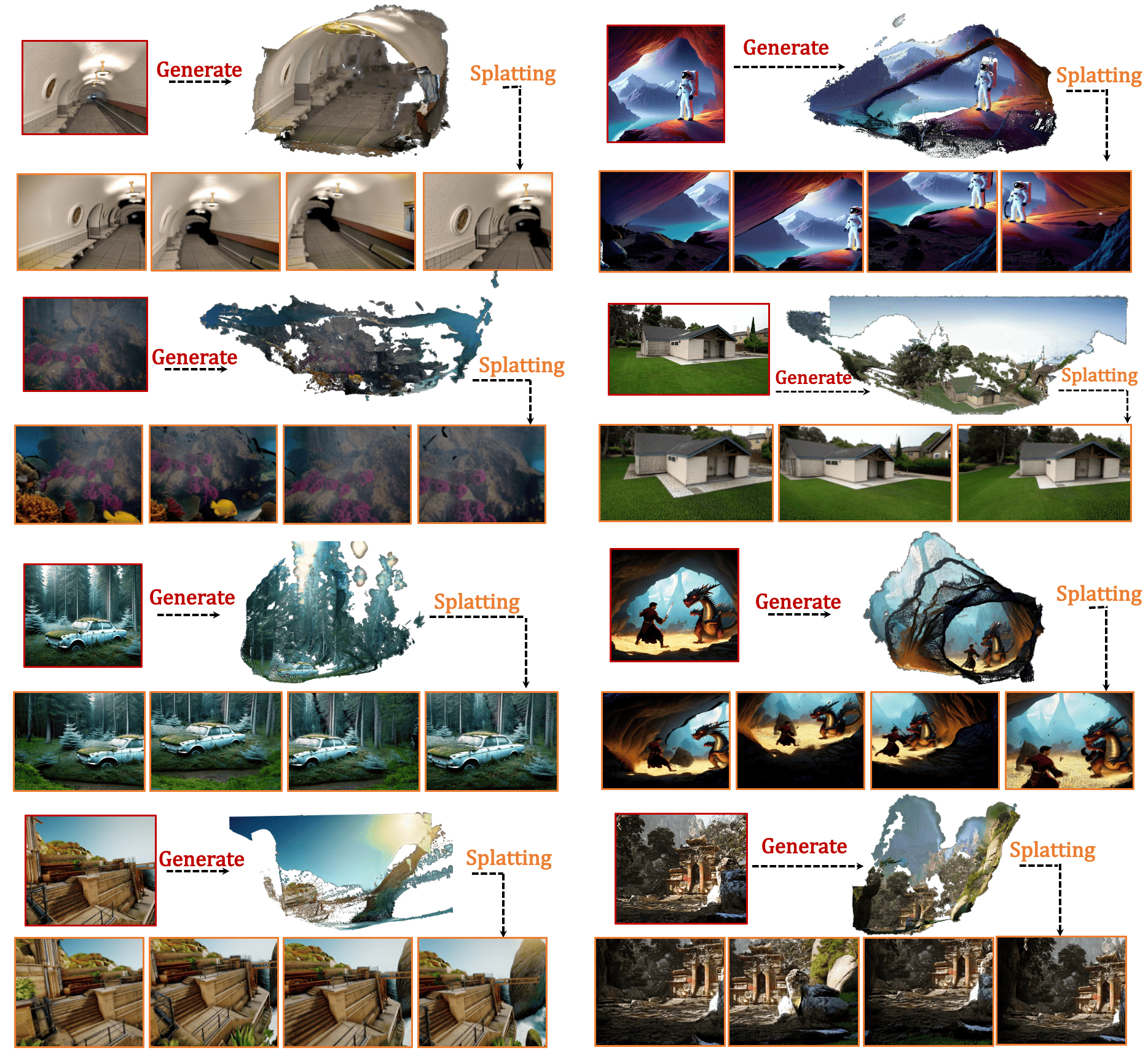}
\end{center}
\vspace{-5pt}
\caption{ \textit{Image-to-3D scenes.}
In each example, VistaDream generates a 3D Gaussian field based on the input image (red box), which is capable of rendering novel view images (orange box)
}
\label{fig:supp}
\end{figure*}

\begin{figure*}
\begin{center}
\includegraphics[width=\linewidth]{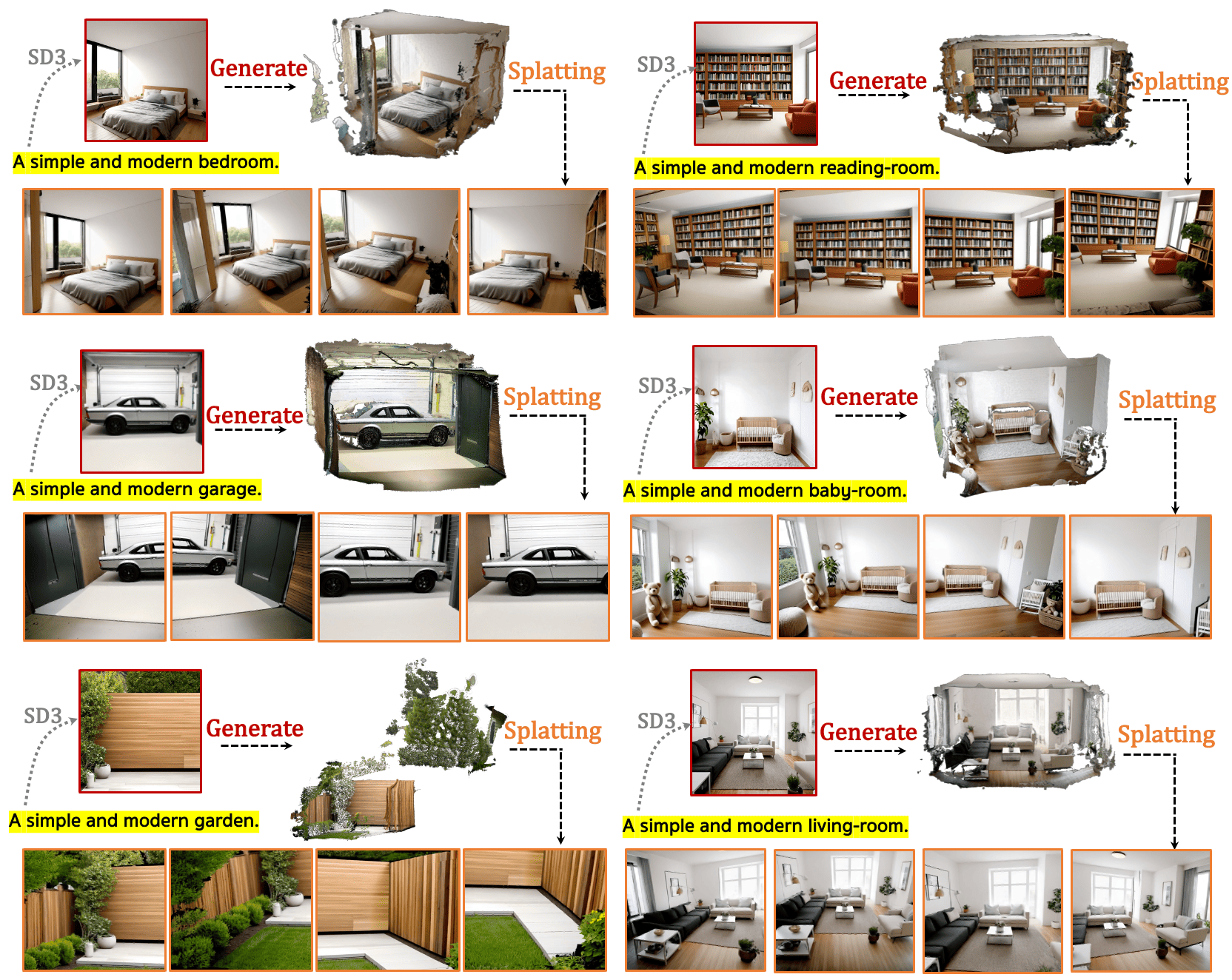}
\end{center}
\vspace{-5pt}
\caption{ 
\textit{Text-to-3D scenes.}
In each example, we use Stable Diffusion 3 to generate an image based on the input text (marked in yellow). Subsequently, VistaDream generates a 3D Gaussian field from the input image (red box), which can be used to render novel view images (orange box).
}
\label{fig:text}
\end{figure*}

\begin{figure*}
\begin{center}
\includegraphics[width=\linewidth]{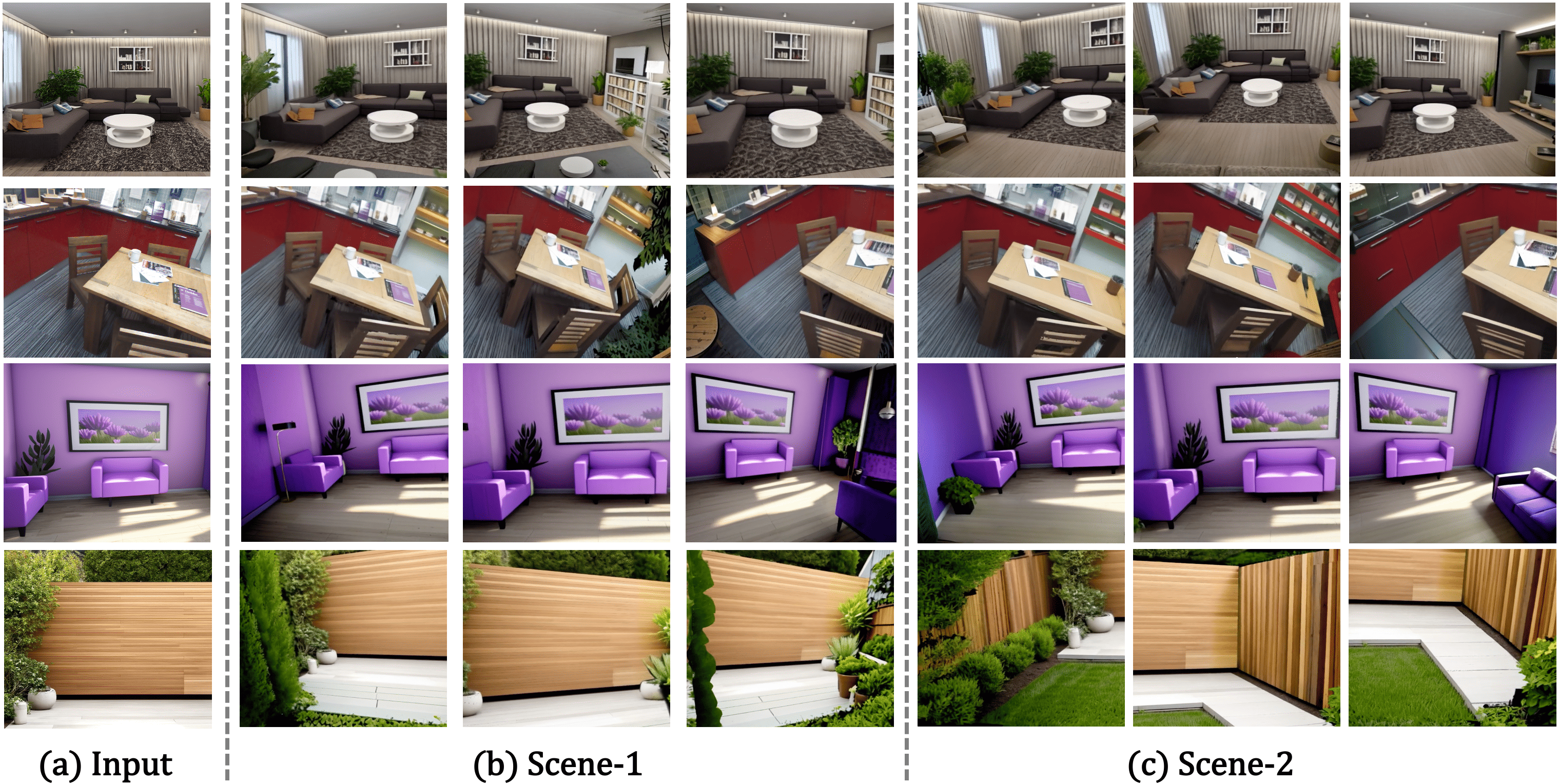}
\end{center}
\vspace{-5pt}
\caption{ 
\textit{Different plausible scenes generated by VistaDream from the same input image.}
}
\label{fig:multi}
\end{figure*}

Given various styles of input images, the results in Fig.~\ref{fig:main} and Fig.~\ref{fig:supp} demonstrate that VistaDream produces clear, accurate, and highly consistent 3D scenes. In Fig.~\ref{fig:text}, VistaDream achieves scene reconstruction from text inputs by incorporating a text-to-image generation model~\citep{sd3}. 
Moreover, in Fig.~\ref{fig:multi}, for the same input image, our method can generate different plausible scenes using different random seeds.
More videos and interactive demos are provided in the supplementary materials.

\subsection{Failure cases}
\begin{figure*}
\begin{center}
\includegraphics[width=\linewidth]{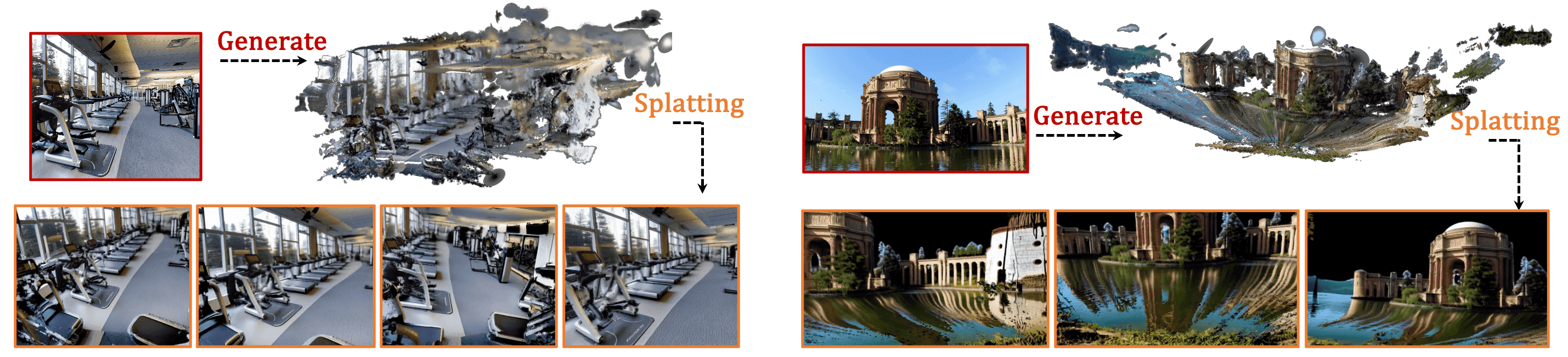}
\end{center}
\vspace{-5pt}
\caption{ 
\textit{Typical failure cases.}
The nearby objects contain significant distortion due to the inaccurate depth estimation of GeoWizard~\cite{geowizard}.
}
\label{fig:fail}
\end{figure*}

In Fig.~\ref{fig:fail}, we present two typical failure cases where distortion occurs in nearby objects. This is due to the inaccurate depth estimation from the monocular depth estimator like GeoWizard~\cite{geowizard}, particularly for objects near to the camera. Improving the quality of depth estimation may solve these issues, which we leave as future work.

\end{document}